\documentclass[letterpaper, 10 pt, conference]{ieeeconf}  

\IEEEoverridecommandlockouts                              

\usepackage{graphics} 
\usepackage{graphicx}
\usepackage{epsfig} 
\usepackage{times} 
\usepackage{amsmath} 
\usepackage{amssymb}  
\usepackage[font=small]{caption}
\usepackage{subcaption}

\usepackage{fontenc}
\usepackage{mathtools}
\usepackage{color}
\usepackage{citesort}
\usepackage[bottom]{footmisc}
\usepackage[ruled,vlined]{algorithm2e}
\usepackage{epstopdf}
\usepackage{tabularx}
\usepackage{hhline}
\usepackage{url}

\definecolor{grey_color}{RGB}{84,84,84}

\DeclareMathAlphabet{\pazocal}{OMS}{zplm}{m}{n}

\title{\LARGE \bf
Aerial Drop of Robots and Sensors for Optimal Area Coverage
}

\author{Kostas Alexis
}

\begin{document}

\maketitle
\thispagestyle{empty}
\pagestyle{empty}

\begin{abstract}
The problem of rapid optimal coverage through the distribution a team of robots or static sensors via means of aerial drop is the topic of this work. Considering a nonholonomic (fixed--wing) aerial robot that corresponds to the carrier of a set of small holonomic (rotorcraft) aerial robots as well as static modules that are all equipped with a camera sensor, we address the problem of selecting optimal aerial drop times and configurations while the motion capabilities of the small aerial robots are also exploited to further survey their area of responsibility until they hit the ground. The overall solution framework consists of lightweight path--planning algorithms that can run on virtually any processing unit that might be available on--board. Evaluation studies in simulation as well as a set of elementary experiments that prove the validity of important assumptions illustrate the potential of the approach.
\end{abstract}

\section{Introduction}\label{sec:intro}

Area coverage by means of a multi--robot team or via optimal distribution of sensors is a problem that has been vastly investigated. The research community has proposed highly elaborated results that can guarantee optimal coverage and autonomous exploration~\cite{SIP_AURO_2015,APST_MSC_2015,bircher_robotica,BABOOMS_ICRA_15,papachristos2016distributed,NBVP_ICRA_16,bircher2016receding,galceran2013survey,hover2012advanced,cortes2004coverage,englot2012sampling,oettershagen2016long,papachristos2016augmented,papachristos2017autonomous,balta2017integrated} while recently a set of inspection path--planning algorithms were proposed and experimentally evaluated using aerial vehicles. However, as long as large areas that have to be inspected from the air subject to hard time constraints, then the endurance limitations of small aerial robots make such missions almost impossible. To overcome this problem, the idea of aerial dropping of a set of small robots or even static sensing modules is considered. In such a scenario, a relatively large carrier aerial robot -- typically a fixed--wing system~\cite{oettershagen2016long,oettershagen2014explicit} characterized by significant payload and endurance -- can be used to distribute efficiently and at meaningful spots the multi--systems (robots and static sensors) team while the motion capabilities of the small aerial robots~\cite{alexis2012model,alexis2014robust,huerzeler2013configurable} to be dropped can be further exploited in order to more thoroughly survey the area until the moment their batteries get depleated. 

This work aims to address some of the technical challenges of this particular idea of optimal aerial coverage using a team of aerially-dropped robots and static sensors. We draw inspiration by existing methods in the field of aerial coverage and rely on proven and experimentally verified characteristics of control laws that are widely used within the autopilots of aerial robots with typical configurations (e.g. multicopters). More specifically, we benefit from the fact that computationally very lightweight controllers such as the family of geometric controllers~\cite{lee2010geometric,fernando2011robust,lee2010control,lee2013nonlinear,goodarzi2013geometric,lee2014geometric} present \textit{almost} global stability characteristics which essentially allows us to consider that we can drop a small multicopter from the air and recover from \textit{almost} any attitude configuration. 

This manuscript is structured as follows. Section~\ref{sec:team} overviews the considered multiple classes of systems team, followed by Section~\ref{sec:method} overviewing the coverage path planning via aerial drop. Evaluation studies are presented in Section~\ref{sec:evaluation}, while conclusions are drawn in Section~\ref{sec:Concl}.

\section{The Multi--Systems Team}\label{sec:team}

The envisaged operational capacity relies on a multi--systems team that comes in the form of multiple micro aerial or ground vehicles as well as miniature static sensors that can alltogether be transported and selectively released by a larger carrier fixed--wing aerial robot. The concept of an unmanned aerial vehicle carrying and selectively releasing other subsystems is not new: air--to--surface missiles capable of releasing smaller bombs over a predefined trajectory have been availabe for several years. Within the framework of this work, we extend this concept to the problem of rapid area coverage and long--term area monitoring using a combination of micro robots, sensors and a carrier vehicle. To address this challenge, each of the micro robots and sensors is considered to be equipped at least with a micro--HD camera and an Inertial Measurement Unit (IMU), the micro--aerial vehicles are considered to be able of attitude control and absolute heading estimation, the micro ground robots are thought to be able of waypoint navigation which is also the case for the carrier fixed--wing aerial robot. All the subcomponents are further considered to be equipped with a wireless communications module with sufficient range and bandwidth capabilities to transmit largely downsampled images along with multi--hop capacity. These considerations are perfectly reasonable if one considers the current state--of--the--art that provides extremely miniaturized camera systems, low--cost IMUs and open--source autopilots that provide means of attitude control and waypoint navigation for both aerial and ground robots~\cite{PixhawkWebsite}. 

\section{Coverage Via Aerial Drop Path--Planning}\label{sec:method}

Coverage using the envisaged robotic team allows rapid exploration, mapping and long--term monitoring of remote hostile areas without putting any important equipment at any major risk. With the carrier fixed--wing aerial robot flying at high--altitudes, this only relatively more expensive equipment is largely sacured. On the contrary as long as a very rough awareness of the environment has been gained, the micro robots and sensors are released in order to derive and reconstruct a denser map of the environment and establish a network of mobile communications within the whole area of interest. The envisaged application scenarios contain those of rapid exploration for search--and--rescue in hazzardous and radioactive environments (e.g. after the unfurtunate events of chemical or nuclear explosion) as well as rapid invasion for area monitoring for defense operations. 

\begin{figure}[htbp]
\centering
\includegraphics[width=0.95\columnwidth]{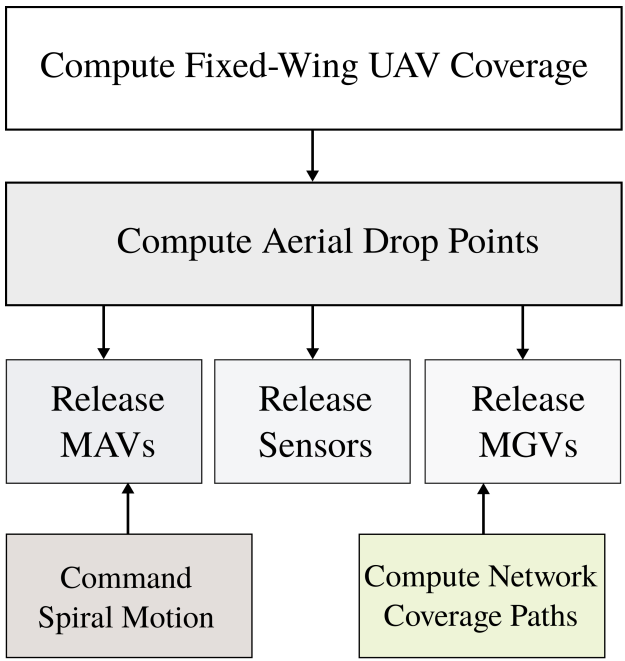}
\caption{Main algorithmical components of the proposed area coverage via means of autonomous aerial dropping and multi--robot collaboration.}
\label{fig:AerialDropPathPlanning}
\end{figure}

To enable this multi--systems team to operate, a delicate, efficient and computationally lightweight path planning framework has to be established. Towards this goal, the overall pipeline depicted in Figure~\ref{fig:AerialDropPathPlanning}. Essentially, the designed framework consists of the intelligent combination of a series of subcomponents, from efficient coverage methods, to basic motion primities, concepts of optimized robot and sensor distribution for network communication coverage and more. It is to be highlighted, that each algorithm proposed to be executed within this framework is tailored to the expected computational and sensorial capabilities of each robot. In that sense, the carrier fixed--wing UAV is expected to be able to execute autonomous navigation through predefined GPS points, which is also the case for the MGVs (once they have touched the ground), while the aerial--dropped MAVs are considered unable to track a specific trajectory during their fall. Within the following subsections, the role of each individual of these subcomponents will be explained and detailed. 

\subsection{Motion Models}

For each robot, a motion model is considered to rule and be able to describe its motion. More specifically, the carrier fixed--wing UAV motion is considered to be described with the formulation of Dubins Airplane paths~\cite{owen2015implementing}, the MAVs are considered to be able to fly holonomic paths, the MGVs are considered to perform Dubins car curves, while naturally the static sensors employ no motion model. As holonomic paths are straightforward and the Dubins car model is well know, within this work, only the Dubins Airplane model is briefly overviewed while an open--source tool is provided and can be found online at~\cite{DubinsAirplaneRepo}.

\subsection{Initial Rapid Area Coverage}

The initial area coverage is achieved using the fixed--wing UAV, flying with high--speed above the overall area to be mapped. The purpose of this initial flight is to gain a first knowledge that could be potentially useful for re--planning purposes but also to collect a first set of aerial images from a close--to--level platform that will become the basis of all map reconstructions. To deploy an efficient area coverage pipeline we split the problem into that of finding a small set of viewpoints that guarantees full coverage from a given altitude and subsequently compute the optimal, shortest--flight path length, tour among them. Within such a framework, the required amount of evenly distributed viewpoints is first computed based on a geometric model of a nadir--facing camera on the UAV, with a Field of View (FoV) $\phi$ flying at $z^c$. With an area with $xy$ dimenstions $[d_x,d_y]$, the amount of viewpoints is:

\begin{eqnarray}
 N_v &=& \frac{d_x d_y}{\Delta_c^2},~\Delta_c = 2 z^c \tan(\phi /2 )
\end{eqnarray}
With this being the required amount of evenly distributed viewpoints in order to ensure full coverage, the Lloyd's algorithm (\textit{Voronoi iteration}~\cite{du1999centroidal,du2006convergence,cortes2004coverage} is employed to find them. The Lloyd's algorithm is capable of finding evenly--spaced sets of points in subsets of Euclidean spaces and partitions of these subsets into uniformly sized convex cells. The Lloyds algorithm repeatedly finds the centroid of each set in the partition and subsequently repartitions the input according to which of these centroids is the closets. The partitioning procedure is based on Voronoi diagrams. The algorithm starts by a random initial placement of the $N_v$ points in the $xy--\textrm{space}$ and it then iteratively runs by alternating within the following two steps:

\noindent \textbf{\textsc{Assignment Step:}} Assign each point to the cluster whose mean yields the least within--cluster sum of squeares. Since the sum of squares in the squared Euclidean distance, this is intuitively the ``nearest'' mean. Equivalently, this process generates the voronoi diagram generated by the meansL 

\begin{eqnarray}
 S_i ^{k} = \{ x_p : ||x_p - m_i^k ||^2 \le ||x_p - m_j^k ||^2,~\forall j,~1 \le j \le n \}
\end{eqnarray}
where each $x_p$ point coordinates is assigned to exactly one $S_i^k$ (Voronoi) cluster, $m_i, m_j$ stand for the mean positions and $k$ is the current algorithm iteration step.

\noindent \textbf{\textsc{Update Step:}} Calculate the new means to be the centroids of the points in the new (Voronoi) clusters:

\begin{eqnarray}
 m_i ^{k+1} = \frac{1}{|S_i^k|_{x_j \in S_i ^k} } \sum x_j 
\end{eqnarray}
where it has to be pointed that since the arithmetic mean is a least--squares estimator, this also minimizes the within--cluster sum of squares objective.

Through this process, at each iteration step the points arrive in a more even distribution with closely spaced points moving further apart and widely spaced points moving closer together. The algorithm convergence is rather slow and therefore to enable practical solution times, a practical threshold on the rate of improvement is imposed to step the execution once a ``good enough'' \textit{almost--}even distribution is found. 

Knowing the points that have to be visited, the algorithm proceeds into the computation of the optimal tour among them. This essentially means that the assymetric Traveling Salesman Problem (TSP) based on the assymetric costs of traveling point from each point $A$ to each other point $B$ using the Dubins Airplane paths model has to be solved. Regarding the computation of paths from one point to the other, the Dubins Airplane model in~\cite{owen2015implementing} describe $16$--different cases of paths, all composed from the combinations of three or more right or left turns (R/L) with minimum turning radius $R_{\min}$ and straight lines (L) with or without ascending along the maximum flight path angle $\gamma_{\max}$. The employed Dubins Airplane model explicitly accounts for low--altitude, medium--altitude and high--altitude change between the origin $\mathbf{q}_0 = [x_0,y_0,z_0,\psi_0]^T$ and the destination $\mathbf{q}_F = [x_F,y_F,z_F,\psi_F]^T$. The thorough description of the Dubins Airplane model is not within the scope of this work, however an open--source code release can be found at~\cite{DubinsAirplaneRepo}. Once, all possible Dubins Airplane connections have been computed, their length populates an assymetric cost--function that will be the basis for the solution of the corresponding TSP that will provide the optimal order to visit all observation points. In order to enable fast computation of such a solution, the efficient Lin--Kernighan--Helsgaun (LKH) solving algorithm~\cite{helsgaun2000effective,lin1973effective} is employed. It is known that local search with $k$--change neighborhoods, $k$--opt as it is called, is one of the most widely used heuristic algorithms to solve TSP. $k$--opt belongs to the family of tour improvement algorithms and within each step, $k$ links of the current tour are replaced by $k$ links in a way that a shorter tour is achieved. Although $k$--opt may in some cases take exponential number of iterations, such undesirable effects are very rare in practice. Usually, high--quality solutions are obtained in polynomial time~\cite{helsgaun2000effective,lin1973effective}. Guidelines on how to use and invoke the execution of the LKH--solver can be found at \url{https://github.com/unr-arl/LKH_TSP}

\subsection{Optimal Aerial Dropping}

Selection of the points of aerial drop is essential for this approach for rapid coverage. With a motion--enabled or static sensing agent being dropped from high altitude it is essential that sufficient information is gothered at the very initial moments and definetely before the body has accelerated significantly. Furthermore, sensor components that have appropriately fast dynamics characteristics have to be employed. From a technological perspective this means that if once considers a global shutter camera, then the response characteristics and the shutter mechanisms respond sufficiently fast, which indeed is within the capabilities of nowadays off--the--shelf available modules. 

With the goal of gathering sufficient information at the initial phase of the aerial drop and considering a conic camera sensor with field--of--view $\phi$ and a nadir--mounted configuration, the dropping points are selected to be in positions and altitudes that overall already guarantee full coverage from the first collected samples. To compute these points effectively and distribute them in space in an efficient way in terms of path--planning they are distributed in the $xy$--plane via the utilization of the Lloyd's algorithm (\textit{Voronoi iteration})~\cite{du1999centroidal,du2006convergence,cortes2004coverage} that guarantees even distribution. Subsequently, the altitude that is associated to each dropping point is computed such that the camera cone provides full coverage of the corresponding Voronoi cell. Note that due to the expected variation on the attitude of each sensing agent, overlap with the images collected by the other agents will be definetely achieved. 

Finally, once all aerial drop points have been computed, a combination of the LKH--TSP solver with Dubins Airplane--computed point--to--point paths is employed in order to derive the optimal tour such that the fixed--wing carrier UAV visits all the aerial drop points at minimum time. 

\subsection{MAV Motion Primitive}

Once dropped, the motion--enabled sensing agents (MAVs with a camera) are commanded to follow a motion primitive that improves the expectancy to cover all the desired area that belongs to their allocated Voronoi cell. The decision on the motion primitive was based on the necessity to be meaningfully executable without a global frame or reference while still providing multiple views from different positions and orientations. A simple coverage pattern has to be selected given the mission approach and the type of very small systems. In particular, a spiral--like based on the harmonic variation of the roll and pitch angles was selected as the MAV motion primitive. This specifically consists of the following roll and pitch commands:

\begin{eqnarray}
 \phi^r &=& A(t)\sin(\omega t) \\
 \theta^r &=& A(t)\cos(\omega t)
\end{eqnarray}
where $\omega$ should be chosen in correlation with the altitude and the size of the Voronoi cell and a heuristic formula $\omega_ = 5.2 S_V/\Delta z$ ($S_v, \Delta z$ the Voronoi cell area and the aerial drop altitude respectively) was found to be applicable. Subject to these commands, an almost harmonic motion is executed (considering fixed heading angle) while the altitude component relies on the thrust levels which might be chosen according to the expected battery endurance. In general an as constant and as slow falling velocity as possible is desired. Figure~\ref{fig:MAVmotionpattern} illustrates an example of this motion pattern:

\begin{figure}[htbp]
\centering
\includegraphics[width=0.95\columnwidth]{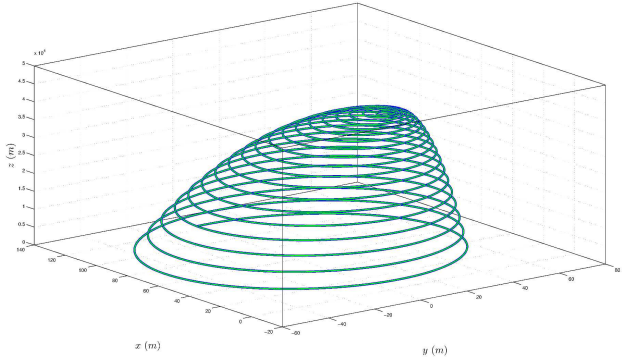}
\caption{Example of possible -conceptual- MAV motion pattern for superior aerial drop coverage}
\label{fig:MAVmotionpattern}
\end{figure}

\subsection{MGV Communication Coverage}

\begin{figure}[htbp]
\centering
\includegraphics[width=0.95\columnwidth]{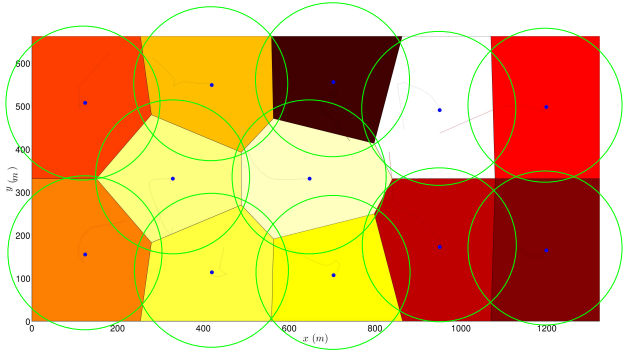}
\caption{Example of the MGV spatial distribution for communication coverage}
\label{fig:MGVdistribution}
\end{figure}

A subset of the aerially--dropped moving agents can possibly be MGVs. These vehicles, once landed should execute an optimal communication coverage spatial distribution algorithm in order to increase the confidence that a solid communication channel gets established between all sensors and guarantees that all the data are collected and transmitted to the human--operators or the carrier fixed--wing UAV. Under the assumption that each MGV is equipped with GPS, then after a Geodetic WGS84 to local planar earth ENU coordinates transformation, the centroids of the Voronoi tesselation can be precomputed in order to ensure spatial distribution uniformity of the active communicating agents and the MGVs can drive themselves from their final points that they landed. The steering is based on the well--known Dubins car model while the allocation of which MGV goes to which Voronoi centroid is decided based on enumaration to find the optimal case of overall minimum additive distances from all the MGVs to all the Voronoi centroids. Figure~\ref{fig:MGVdistribution} presents an example of such a uniform spatial distribution. 

\section{Evaluation Studies}\label{sec:evaluation}

\begin{figure}[htbp]
\centering
\includegraphics[width=0.99\columnwidth]{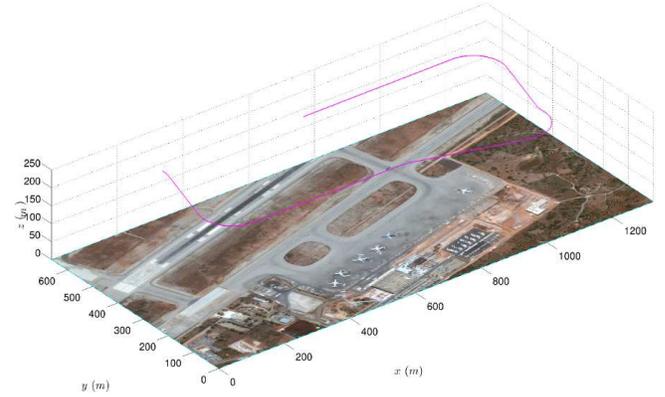}
\caption{Rapid high--altitude coverage path using the carrier fixed--wing vehicle in order to gather basic intelligence and collect a basic set of nadir--looking aerial views that will become the basis of the area reconstruction .}
\label{fig:FWcov}
\end{figure}

To evaluate the proposed aerial drop--based rapid area coverage strategy, a large set of simulation cases were considered. Different agents teams were considered consisting by the combination of aerial robots with non--actuated sensors as well as teams that also contained MGVs. The considered scenario is that of a Search--and--Rescue mission taking place at an airport in the world. For all cases, an initial rapid high--altitude coverage path is conducted by the fixed--wing UAV in order to provide rapid intelligence but also collect a key set of nadir--looking aerial views that will correspond to the basis of the area reconstruction. The computed path for the area of an airport is shown in Figure~\ref{fig:FWcov}.


\begin{figure}[htbp]
\centering
\includegraphics[width=0.99\columnwidth]{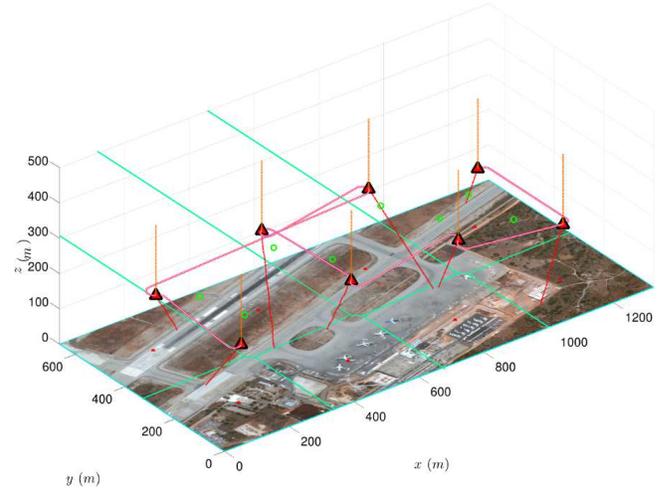}
\caption{Rapid aerial drop--based area coverage using $4$ MAVs and $4$ static sensors released by the fixed--wing carrier vehicle.}
\label{fig:4b4}
\end{figure}

\begin{figure}[htbp]
\centering
\includegraphics[width=0.99\columnwidth]{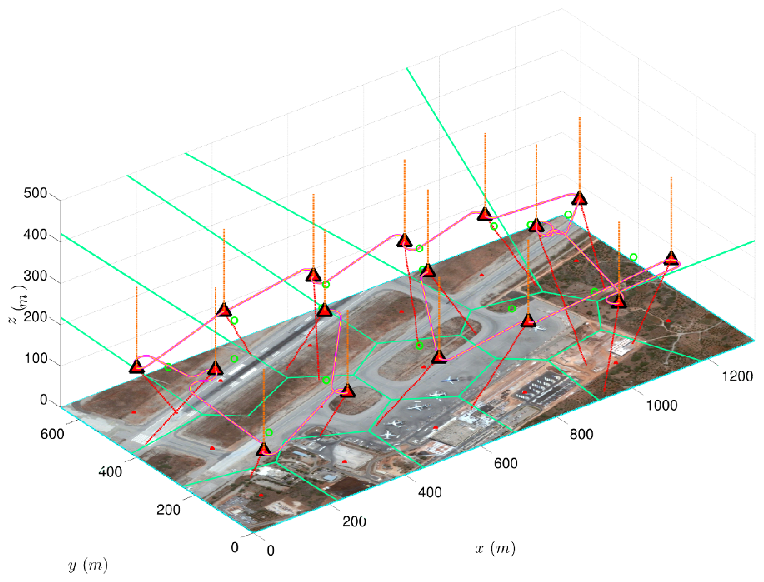}
\caption{Rapid aerial drop--based area coverage using $8$ MAVs and $8$ static sensors released by the fixed--wing carrier vehicle.}
\label{fig:8b8}
\end{figure}

\begin{figure}[htbp]
\centering
\includegraphics[width=0.99\columnwidth]{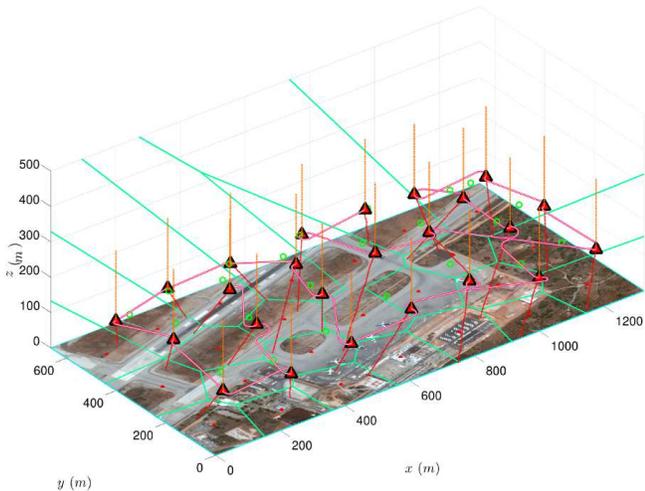}
\caption{Rapid aerial drop--based area coverage using $12$ MAVs and $12$ static sensors released by the fixed--wing carrier vehicle.}
\label{fig:12b12}
\end{figure}

Initially, the case of 4 MAVs and 4 static sensors, 8 MAVs and 8 static sensors as well as 12 MAVs and 12 static sensors is considered. The corresponding results are shown in Figures~\ref{fig:4b4},~\ref{fig:8b8},~\ref{fig:12b12}. As becomes evident, the larger the amount of agents to be released, the smaller the area each of them has to cover. Furthermore, it is shown that the proposed aerial drop path computation algorithm handles the nonholonomic constraints of the carrier fixed--wing vehicle while the aerial drop points are the optimized in terms of area coverage distribution. In addition it is depicted that as long as the aerial drop altitude is small, the MAVs have very limited time to perform the advanced coverage maneuver as the vertical speed becomes soon very fast while their planar velocities are in general small. 

\begin{figure}[htbp]
\centering
\includegraphics[width=0.99\columnwidth]{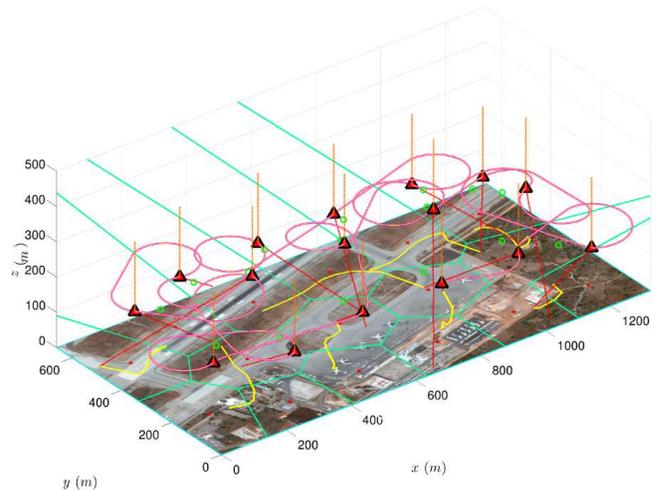}
\caption{Rapid aerial drop--based area coverage using $8$ MAVs and $8$ MGVs released by the fixed--wing carrier vehicle. Once the MGVs get deployed on the ground, they move towards achieving optimal communications--based coverage. Considering that the communications range of the MGVs is sufficient, then the MAVs and the static sensor are able to communicate through that network and also robustify and improve it. }
\label{fig:8b8b0}
\end{figure}

\begin{figure}[htbp]
\centering
\includegraphics[width=0.99\columnwidth]{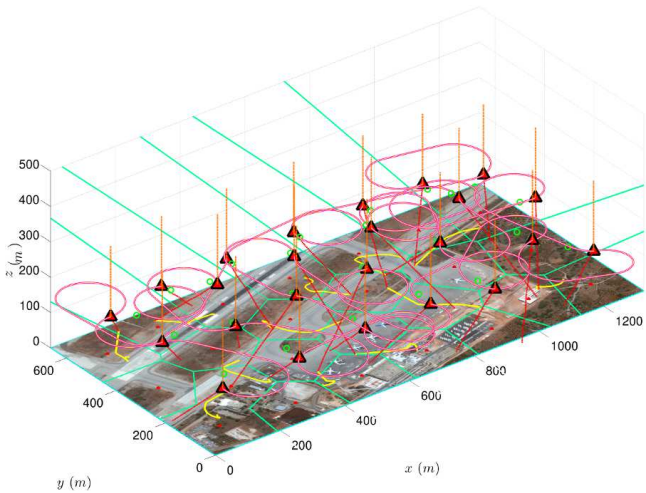}
\caption{Rapid aerial drop--based area coverage using $12$ MAVs and $12$ MGVs released by the fixed--wing carrier vehicle. Once the MGVs get deployed on the ground, they move towards achieving optimal communications--based coverage. Considering that the communications range of the MGVs is sufficient, then the MAVs and the static sensor are able to communicate through that network and also robustify and improve it.}
\label{fig:12b12b0}
\end{figure}

Subsequently, the case of advanced robotic teams that consisting of MAVs and MGVs in order to achieve superior communication coverage on the ground was considered. As shown, the MGVs act like static sensors during the aerial drop phase, while subsequently execute the optimal communications--coverage path. Similarly, Figure~\ref{fig:8b8b0} presents the case of 8 MAVs and 8 MGVs without any static sensor while Figure~\ref{fig:12b12b0} presents the case of 12 MAVs, 12 MGVs and no static sensor. As shown, the use of MGVs employed with sufficient communication range, ensures that complete communications coverage is achieved on the ground, while the deployed MAVs are not only able to communicate via the MGVs--established network but they also robustify and improve it.


The aforementioned test--cases stand as illustrative scenarios that present the capability of the proposed pipeline to compute an initial high--altitude coverage path and subsequently determine the aerial drop points and the path that steers the fixed--wing carrier vehicle optimally among them. Furthermore, a distributed algorithm is employed to guide the MGVs for optimal communications--based area coverage as long as these vehicles are considered.

\section{Conclussions}\label{sec:Concl}

A strategy for rapid area coverage based on aerial dropping of MAVs, MGVs as well as static sensors was proposed. The carrier fixed--wing UAV not only releases the agents but also executes an initial coverage path to gather intelligence and collect a key set of aerial views that will then be used to combine the images from the several agents and compute an overall consistent combined map. Based on the optimally computed aerial drop points, an optimized tour for the nonholonomic carrier fixed--wing UAV is computed and the heterogeneous team of agents (MAVs, MGVs, static sensors) is considered to be released. Due to the nature of the falling path, the aerial views have sufficient overlap in multiple cases. Furthermore, the proposed approach employs the dropped MGVs with the capacity to guide and steer themselves towards achieving optimal coverage based on their communication ranges.


\end{document}